\documentclass[10pt, conference, compsocconf]{IEEEtran}

\ifCLASSINFOpdf
\else
\fi
\usepackage[cmex10]{amsmath}
\usepackage{graphicx} 

\usepackage{algorithmic}

\usepackage{array}

\usepackage{mdwmath}
\usepackage{mdwtab}

\usepackage{eqparbox}

\usepackage[font=footnotesize]{subfig}

\usepackage{stfloats}

\usepackage{url}

\graphicspath{{images/}}
\usepackage{amssymb}
\DeclareMathOperator*{\argmax}{arg\,max}

\DeclareMathOperator*{\argmin}{arg\,min}
\begin{document}
%
\title{Active Learning from Positive and Unlabeled Data}




%
\author{\IEEEauthorblockN{Alireza Ghasemi\IEEEauthorrefmark{1},
Hamid R. Rabiee\IEEEauthorrefmark{1},
Mohsen Fadaee\IEEEauthorrefmark{1}, 
Mohammad T. Manzuri\IEEEauthorrefmark{1} and
Mohammad H. Rohban\IEEEauthorrefmark{1}}
\IEEEauthorblockA{\IEEEauthorrefmark{1}Digital Media Lab, AICTC Research Center\\
Department of Computer Engineering\\
Sharif University of Technology\\
Azadi Avenue, Tehran, Iran\\
a\_ghasemi@ce.sharif.edu,rabiee@sharif.edu,m\_fadaee@ce.sharif.edu,manzuri@sharif.edu,rahban@ce.sharif.edu}
}


\maketitle

\begin{abstract}

During recent years, active learning has evolved into a popular paradigm for utilizing user's feedback to improve accuracy of learning algorithms. Active learning works by selecting the most informative sample among unlabeled data and querying the label of that point from user. Many different methods such as uncertainty sampling and minimum risk sampling have been utilized to select the most informative sample in active learning. Although many active learning algorithms have been proposed so far, most of them work with binary or multi-class classification problems and therefore can not be applied to problems in which only samples from one class as well as a set of unlabeled data are available.

Such problems arise in many real-world situations and are known as the problem of learning from positive and unlabeled data. In this paper we propose an active learning algorithm that can work when only samples of one class as well as a set of unlabelled data are available. Our method works by separately estimating probability desnity of positive and unlabeled points and then computing expected value of informativeness to get rid of a hyper-parameter and have a better measure of informativeness./
Experiments and empirical analysis show promising results compared to other similar methods.

\end{abstract}

\begin{IEEEkeywords}
active learning; one-class learning; learning from positive and unlabelled data; semi-supervised learning;uncertainty sampling

\end{IEEEkeywords}

%
\IEEEpeerreviewmaketitle

\section{Introduction}
There has been an explosive growth in the amount of available digital content during recent years. This is mostly because of the technological development and evolution of web which causes vast amount of digital media to be created. Managing and organizing such a large amount of data is a tedious task which has to be automated as much as possible. To facilitate this task, machine learning algorithms have been utilized for automatic classification of digital objects. However, achieving acceptable classification accuracy by machine learning algorithms requires large amount of labelled data to be used for training the algorithms. Labeling data has to be done manually and therefore it is a time consuming and expensive task by itself. Because of this, methods which try to exploit unlabeled data to improve accuracy of classification have been of prime interest in recent years. One of such methods is active learning which tries to improve classification accuracy by posing a limited number of queries to a user who can predict label of unlabelled data.

Active learning works by selecting among unlabeled data, the most informative data sample. The informativeness of a sample is the amount of accuracy gain achieved after adding it to the training set. Many paradigms have been proposed to asses informativeness of data samples for active learning. One of the popular approaches is selecting the most uncertain data sample, i.e the data sample in which current classifier is least confident. Some other approaches are selecting the sample which yields a model with minimum risk or the data sample which yields fastest convergence in gradient based methods \cite{settles1}.

Although many active learning algorithms have been proposed in the literature so far, most of them require that the training set contain labeled data of all classes, or at least two of them. In other words, most active learning algorithms work with binary or multi-class classification problems and therefore can not be applied to problem in which only samples from one class and a set of unlabeled data are available. Such problems arise in many real-world situations like information retrieval, document classification and ranking and are knows as the problem of learning from positive and unlabeled data. Although a wide range methods have been proposed for learning from positive and unlabeled data \cite{lfpu}, few efforts have been made to propose active learning methods consistent with this settings.

In this paper, we propose an uncertainty-based active learning algorithm which requires only samples of one class and a set of unlabeled data in order to operate. The principal contribution of our work is twofold: First, we use Bayes' rule and density estimation to avoid the need to have a model of all classes for computing the uncertainty measure. This allows us to use popular uncertainty measures of active learning while utilizing only positive and unlabeled samples, rather than using both positive and negative data. Our second contribution is that we define the prior probability of positive class as a random variable and compute expected value of the uncertainty measure by integrating over all possible parameter values. This technique reduces the number of input parameters of the problem.

At the rest of this paper, we first review recent related works in the fields of active learning and active one-class learning (section \ref{review}). Then, we propose the two approaches to measure uncertainty, utilizing only positive and unlabeled data (section \ref{prop}). After that, we explain our testing framework, analyze experimental results of the proposed methods, and compare them with other methods of this field (section \ref{test}). Finally, we conclude the paper and present ideas for future improvements of the methods.

\section{Related Works}
\label{review}

Many active learning algorithms have been proposed in the literature so far. \cite{settles1} is a comprehensive survey of recent works in this field.
Among the earliest and most popular active learning paradigms is the uncertainty sampling approach which is based on selecting the least confident sample for querying. The definition of confidence depends on the base classifier in use. For example, \cite{ong1} proposes an active learning approach for SVM which selects for querying the sample which is closest to the separating hyperplane. selecting the sample with minimum margin \cite{margin} and the data sample with maximum entropy \cite{entropy} are other approaches which have been applied to active learning problems.

For classifiers that are unable to define a similarity measure over their predictions, committee-based active learning methods have been proposed. these methods form an ensemble or committee of diverse classifiers and measure uncertainty by the amount of disagreement between committee members' votes for a data sample \cite{settles1}.

In problems where samples of only one class are available, traditional uncertainty assessment methods can not work since they require information about at least two of the classes or their separating hyperplane . Therefore, specific active learning methods are required for one-class problems. One of the earlier works is \cite{aod} which uses active learning for outlier detection. This methods works in two stages: First, a number of unlabeled are selected as negative samples by means of statical methods. Then a traditional committee based active learning algorithm is used to perform active learning on the rest of samples. The main advantage of this approach is that it's flexible and can utilize a wide range of traditional active learning algorithms. 
However, \cite{aod} approaches the problem of one-class learning by a traditional binary classification method. This causes degradation in accuracy of the resulting classifier since the two classes have very different characteristics and should not be treated equally. moreover, because of using two learning algorithms, the runtime complexity of this approach is much higher than other similar methods..
 
Another method for active learning from positive and unlabeled data has been proposed by \cite{gmr}. This paper suggests that the best choice for active learning is selecting the most relevant data sample. the justification behind this claim comes from the nature of relevance feedback in image retrieval applications. In other words, the most informative data will be chosen by the following rule:

\begin{center}
\begin{equation}
\begin{matrix}
x_*=\argmax_{x\in U} f(x)
\end{matrix}
\end{equation}
\end{center}

in which $f(.)$ is the scoring function of one-class learning which is used to rank data samples by their likelihood to the target (positive) class. The main advantages of this method lie in its simplicity and speed. However, since this method does not consider uncertainty in choosing samples, the selected data point may lack informativeness.

A more recent approach has been proposed in \cite{asdd}, which tries to apply active learning to the well-known SVDD method. \cite{asdd} considers likelihood as well as local density of data point to assess their uncertainty. First, the algorithm constructs a neighborhood graph over all data samples. Then, the most informative sample is selected using the following rule:
\begin{center}
\begin{equation}
\label{gornitz}
\begin{matrix}
x_*=\\
\argmin_{x_i\in U} \sigma \frac{||d(x_i,\mathcal{C})-R||}{c} + \frac{1-\sigma}{2k} \Sigma_{x_j\in L\cup U} (\overline{y}_j+1)a_{ij}
\end{matrix}
\end{equation}
\end{center}

In \ref{gornitz}, parameters $c$ and $\sigma$ are used to manipulate the significance of any of two factors in the final decision measure, $d(x_i,\mathcal{C})$ is the distance between $x_i$ and center of sphere formed by the SVDD approach. $R$ is radius of that sphere. $\overline{y}$ is 0 for unlabeled data, $+1$ for positive and $-1$ for negative samples. $a$ is the adjacency matrix of the data neighborhood graph. $a_{ij}=1$ if there is an edge between $x_i$ and $x_j$, and $0$ otherwise.

The main advantage  of \cite{asdd} is that it considers both selection based on uncertainty of data, and exploring unknown regions of the feature space. This fact can be easily inferred from the two terms of equation \ref{gornitz}. 
However, this methods is biased toward exploring regions containing negative data in the feature space.
This causes algorithm to be biased to selecting data which are more likely negative samples.  Due to the nature of one-class learning, positive data are much more valuable than negative data samples and therefore selecting negative samples may not be much helpful in improving classification accuracy. Moreover, constructing the neighborhood graph is a time consuming task and makes the algorithm  infeasible for real-time applications.

\section{The Proposed Approach}
\label{prop}
In this section we present the two proposed approaches for learning from positive and unlabelled data.

\subsection{Expected Margin Sampling}
\label{emss}
Margin sampling is among the well-known approaches used for uncertainty-based active learning in different application domains \cite{settles1}. Margin in this context is defined as the difference between posterior probability of the most likely and the second most likely class. The data sample with smaller margin is intuitively more uncertain and hence more informative for learning the positive class.

The margin sampling strategy for active learning is selecting the data sample from the unlabelled sample pool which has the smallest margin, i.e. the data sample for which two of the classes are as equally likely as possible. In a binary classification problem, the selection rule for margin sampling is:
\begin{center}
\begin{equation}
\label{marsam}
\begin{matrix}
x_*=\argmin_{x\in U} |p(+|x)-p(-|x)|
\end{matrix}
\end{equation}
\end{center}
in which $p(+|x)$ and $p(+|x)$ are posterior probabilities of the two classes respectively. Applying Bayes' rule to \ref{marsam} yields:
\begin{center}
\begin{equation}
\label{bmarsam}
\begin{matrix}
x_*=\argmin_{x\in U} \left|\frac{p(x|+)p(+)}{p(x)}-\frac{p(x|-)p(-)}{p(x)}\right|\\
=\left|\frac{p(x|+)p(+)-p(x|-)[1-p(+)]}{p(x)}\right|
\end{matrix}
\end{equation}
\end{center}
For one-class problems in which there is access to samples of one-class only, $p(x|+)$ can be computed easily from the positive (available) class samples. However, since samples from the other classes (which we call collectively the negative class) are not available for training, $p(x|-)$ can not be estimated directly. Moreover, the $p(+)$ estimate should be computed from other sources of information or according to a priori knowledge about the problem. It may be given as input to the algorithm.

The term $p(x)$ is not usually computed directly in Bayes' rule applications. Since it has a normalizing role in Bayes' formula, it can be set easily such that $p(.|x)$ becomes a probability distribution, i.e. 
\begin{center}
\begin{equation}
\label{px}
\begin{matrix}
p(x)=p(x|+)p(+)+p(x|-)[1-p(+)]
\end{matrix}
\end{equation}
\end{center}
However, when large amount of unlabelled data are available, $p(x)$ can be estimated directly using a parametric or non-parametric \cite{bishop} density estimation approach. Noting the fact that $p(x|+)$ can be estimated from positive samples in the same way that $p(x)$ is estimated from unlabelled data, and the $p(x|-)$ is un-known in one-class settings, the equation can be reorganized to compute an estimate for $p(x|-)$ \cite{denis}:
\begin{center}
\begin{equation}
\label{pxm}
\begin{matrix}
p(x|-)=\frac{p(x)-p(x|+)p(+)}{1-p(+)}
\end{matrix}
\end{equation}
\end{center}
In \ref{pxm}, $p(+)$ is the prior probability of positive class which we assume a priori known. Substituting \ref{pxm} into \ref{bmarsam} yields the following equation for the margin sampling strategy of active learning:
\begin{center}
\begin{equation}
\label{bmarsam2}
\begin{matrix}
x_*=\argmin_{x\in U} \left|\frac{p(x|+)p(+)-\frac{p(x)-p(x|+)p(+)}{1-p(+)}[1-p(+)]}{p(x)}\right|\\
=\left|\frac{p(x|+)p(+)-p(x)+p(x|+)p(+)}{p(x)}\right|
\end{matrix}
\end{equation}
\end{center}
After mathematical simplifications of \ref{bmarsam2} and setting $ a_x=\frac{p(x|+)}{p(x)} $ and $P=p(+)$ to make expressions shorter, we reach the following:
\begin{center}
\begin{equation}
\label{bmarsam3}
\begin{matrix}
x_*=\argmin_{x\in U} \left| 1-2a_xP \right|
\end{matrix}
\end{equation}
\end{center}
In \ref{bmarsam3}, $P$ is assumed a priori known. To relax this assumption and count for uncertainty in $P$, we average over different values of it.

Considering $P$ a random variable, we compute expected value of margin for a data sample, rather than selecting a single value for $P$, i.e.
\begin{center}
\begin{equation}
\label{ems}
\begin{matrix}
x_*=\argmin_{x\in U} \mathbb{E}_P\{\left| 1-2a_xP \right|\}
\end{matrix}
\end{equation}
\end{center}
Since $P$ is a probability value and there is no other prior knowledge available about it, we can assume that values of $P$ come from a continuous uniform distribution. utilizing this fact, computing the expected value is equivalent to integrating over all possible values of $P$:
\begin{center}
\begin{equation}
\label{emsint}
\begin{matrix}
 \mathbb{E}_P\{\left| 1-2a_xP \right|\}=\\
 \int_0^1 \left| 1-2a_xP \right| dP = (1-a_x)sgn(\frac{1}{2}-a_x)
\end{matrix}
\end{equation}
\end{center}
After integration, the resulting form of margin sampling strategy using positive and unlabelled data is as follows:
\begin{center}
\begin{equation}
\label{ems}
\begin{matrix}
x_*=\argmin_{x\in U} \left(1-\frac{p(x|+)}{p(x)}\right)sgn\left(\frac{1}{2}-\frac{p(x|+)}{p(x)}\right)
\end{matrix}
\end{equation}
\end{center}
In which $a_x$ is replaced by its original value $\frac{p(x|+)}{p(x)}$.

To estimate $p(x)$ and $p(x|+)$, we can use any of the parametric or non-parametric density estimation approaches like kernel density estimation or Gaussian mixture density. For computing $p(x|+)$, only positive data samples should be used while for $p(x)$ all data can be utilized.

\begin{figure}[t]
\begin{center}
\includegraphics[width=1in,height=1in]{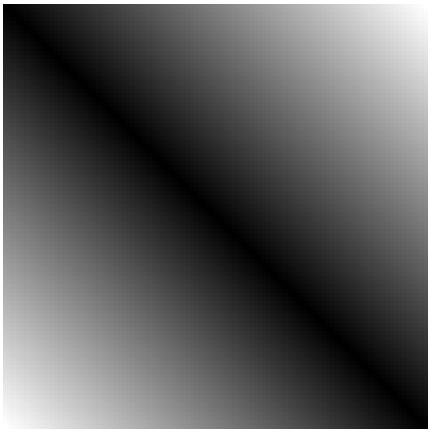}  
  \caption{Data Selection in Margin Sampling}
  \label{msam}
  \end{center}
\end{figure}

Figure \ref{msam} depicts a visualization of margin sampling strategy for two class problems. The horizontal and vertical axis correspond to values of $p(+|x)$ and $p(-|x)$ respectively. The intensity of a point show the margin between posterior probability of the two classes corresponding to the coordinates of the points.
Pseudo-code of the algorithm for active learning from positive and unlabelled data using margin sampling is depicted in figure \ref{alpud}.
\begin{figure}

\begin{algorithmic}[1]
\small
\REQUIRE Set of Positive Target Samples $P$, Set of Negative Outlier Samples $N$
\REQUIRE  Set of Unlabelled Data $U$
\REPEAT 
\STATE $L=P+N$
\STATE $x_*=\argmin_{x\in U} \left(1-\frac{p(x|+)}{p(x)}\right)sgn\left(\frac{1}{2}-\frac{p(x|+)}{p(x)}\right)
$

\STATE Ask label of $x_*$ from user

\IF{$x_*$ is labelled as target by user}

\STATE $P\leftarrow P \cup \{s\}$

\ELSE 

\STATE $N\leftarrow N \cup \{s\}$

\ENDIF

\STATE Perform Learning using the new training set $ L=P+N $.

\UNTIL{Some Stopping Condition is Met}

\end{algorithmic}

\caption{Active Learning from Positive and Unlabelled Data (ALPUD)}
\label{alpud}
\normalsize
\end{figure}

\subsection{Entropy Based Active Learning from Positive and Unlabelled Data}
Another method for assessing uncertainty of data samples in active learning is the well-known entropy method. Entropy is a measure computed for continuous and discrete probability distributions which measures the uncertainty of the distribution. For a discrete distribution $C$, entropy is computed as:
\begin{center}
\begin{equation}
\label{entr}
\begin{matrix}
\mathcal{H}(C)=-\Sigma_{c\in C} p(c)\log(p(c))
\end{matrix}
\end{equation}
\end{center}
Indeed, entropy is a measure of the amount of information included in a probability distribution. the more information a probability distribution carries, the more uncertainty exists in its values.

For active learning tasks, entropy is computed for the posterior class distribution of each data sample. The posterior probability of each class, given the data sample can be considered a discrete probability distribution whose values are class labels. Using this definition and assuming that there are two classes  (positive and negative) in the problem, The entropy-based active learning rule is derived as:
\begin{center}
\begin{equation}
\label{ensam}
\begin{matrix}
x_*=\argmax_{x\in U} \mathcal{H}(.|x)=\\-[p(+|x)\log p(+|x) +p(-|x)\log p(-|x)]
\end{matrix}
\end{equation}
\end{center}
In the same manner as section \ref{emss} and noting the one-class nature of the problem, we can apply Bayes' rule and substitute \ref{pxm} into \ref{ensam} which, after simplification, yielding the following equation for entropy:
\begin{center}
\begin{equation}
\label{ensam2}
\begin{matrix}
\mathcal{H}=-[ a_xP\log(a_xP) + (1-a_xP)\log(1-a_xP)]
\end{matrix}
\end{equation}
\end{center}
Note that we have used  $ a_x=\frac{p(x|+)}{p(x)} $ and $P=p(+)$ again as abbreviations for avoiding long expressions.

Utilizing the same reasoning as in \ref{emss}, we consider $P$ a random variable and compute the expected value of entropy with regard to $P$ instead of computing entropy for single $P$. therefore the entropy-based sample selection rule becomes:
\begin{center}
\begin{equation}
\label{ensam3}
\begin{matrix}
x_*=\\
\argmax_{x\in U} \mathbb{E}_P\{ -[ a_xP\log(a_xP) + (1-a_xP)\log(1-a_xP)]\}
\end{matrix}
\end{equation}
\end{center}
Assuming a continuous uniform distribution over $P$ values, we reach the following rule:
\begin{center}
\begin{equation}
\label{ensam4}
\begin{matrix}
\mathbb{E}_P\{\mathcal{H}\}=\int_0^1  -[ a_xP\log(a_xP) + (1-a_xP)\log(1-a_xP)] dP \\
=\frac{-a_x^2\log(a_x) +a_x +(a_x-1)^2\log(1-a_x)}{2a_x}
\end{matrix}
\end{equation}
\end{center}
Finally, the sample selection strategy for active learning using positive and unlabelled data is derived as:
\begin{center}
\begin{equation}
\label{ensam۵}
\begin{matrix}
x_*=\argmax_{x\in U} \frac{-a_x^2\log(a_x) +a_x +(a_x-1)^2\log(1-a_x)}{2a_x}
\end{matrix}
\end{equation}
\end{center}
In which  $ a_x=\frac{p(x|+)}{p(x)} $. Again, to compute $ p(x)  $ and $ p(x|+) $ we can use any of the parametric or non-parametric density estimation methods.

\begin{figure}[t]
\begin{center}
\includegraphics[width=1in,height=1in]{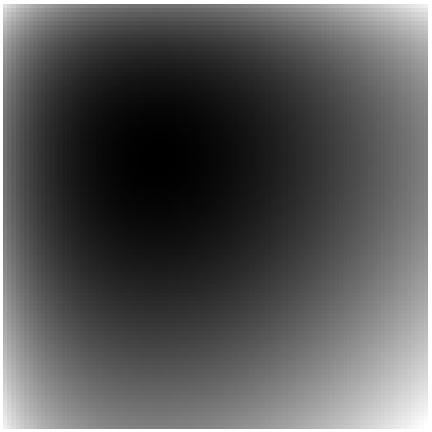}  
  \caption{Data Selection Using Entropy}
  \label{entsam}
  \end{center}
\end{figure}

Figure \ref{entsam} depicts a visualization of margin sampling strategy for two class problems. Algorithm for entropy based active learning is the same as \ref{alpud}, except line 3 which is replaced by \ref{ensam۵}.

\section{Experiments and results}
\label{test}
For evaluation of the proposed methods, we have used various real-world datasets. A number of datasets from the UCI repository \cite{uci} as well as the Caltech image image dataset were used in our experiments for evaluation of the proposed active learning strategies\cite{caltob}. For the Caltech images, CEDD features were extracted from each image \cite{cedd}. Table \ref{dsets} depicts properties of the datasets that have been used in our evaluations.

\begin{table}[t]
\begin{center}
\caption{Datasets Used in Experiments}
\label{dsets}
\begin{tabular}{|l|c|c|}
\hline
{\bf Dataset} & {\bf \# features} & {\bf \# Samples(\# Targets)} \\ \hline
{\bf USPS Digit 3 }& 256 &  9298 (824) \\ \hline
{\bf USPS Digit 4 }& 256 &  9298 (852) \\ \hline
{\bf MNIST Digit 5 }& 784 & 60000 (5421) \\ \hline
{\bf MNIST Digit 6 }& 784 & 60000 (5918) \\ \hline
{\bf ISOLET Class 1 }& 617 & 7279 (300)\\ \hline
{\bf ISOLET Class 2 }& 617 & 7279 (300)\\ \hline
{\bf Caltech  Face Images} & 144 & 3821 (450) \\ \hline
{\bf Caltech  Car Images} & 144 & 3821 (526) \\ \hline
\end{tabular}
\end{center}
\end{table}
\normalsize
\subsection{Evaluation Criteria and Algorithms}

Since in many problems of learning from positive and unlabeled data the goal is indeed a retrieval task, performance measure from the field of information retrieval are popular for evaluation of one-class learning methods \cite{irsvdd}. therefore we used a measure of information retrieval for evaluation of our methods.

There are many performance measures used in the context of information retrieval, the most popular among them are precision and recall\cite{manning}. Here, we have used F1-measure  as the performance measure. It is the harmonic mean of precision and recall which is more meaningful than any of them alone. Using this measure, we penalize situations in which only one of precision and recall has a high value and force the requirement that both of them be within an acceptable range.

Our proposed approaches do not use unique properties of any specific one-class learning algorithms and are independent of the base one-class learner used. Therefore, any one-class learning algorithm can be used as the base classifier. In the experiments, we used SVDD \cite{svdd} as base classifier and kernel density estimation (KDE) as the density estimation method to find likelihood of positive and unlabelled data. SVDD was chosen because of its efficiency and popularity in different applications. KDE was selected because it is a well-known non-parametric density estimation approach and it does not require an expensive training phase.

Although we used SVDD and KDE for the reasons mentioned above, other one-class learning methods (like one-class Gaussian processes \cite{ocgp}) or density estimation methods (like GMM) can be utilized in the algorithm as well.

\subsection{Experiment Setup and Result}
For the experiment setup, initially 200 samples were selected as the pool of unlabelled data, from which samples are selected by the proposed active learning rule for querying. Then, from the remaining data samples,  half of the target samples were selected for training and the other half, in addition to the non-target (outlier) samples was used as the unlabelled data for testing.  All data selections were performed by random sampling.

The Gaussian function was used as th kernel in both KDE and SVDD. The bandwidths $h$ and $\sigma$ as well as other parameters of the experiment were adjusted  by cross-validation. 


The goal is to compare different sample selection strategies. As the baseline method, we used random sample selection, as well as the method proposed in \cite{gmr} and the approach of \cite{asdd}.

We measured difference in F1-measure after adding each sample by any of the proposed and baseline strategies separately and compared them after adding 25 samples. Table \ref{r1} depicts the F1-measure gain after adding 25 samples for any of the mentioned active learning methods. The last two columns of this show the results of our two proposed methods.

\begin{table}[t]
\begin{center}
\caption{Experimental active learning (amount of gain in F1-measure) results after adding 25 samples. The last two columns show the results of our two proposed methods.}
\label{r1}
\tiny
\begin{tabular}{|l|c|c|c|c|c|}

\hline
 & {\bf Random}   & {\bf  \cite{gmr}} & \textbf{\cite{asdd} }&{\bf  ALPUD with Margin}& {\bf  ALPUD with Entropy}\\ \hline
{\bf USPS 3 }&  2.56$\pm 0.29$	& 3.43$\pm 0.09$	& 3.65$\pm 0.08$	& 3.93$\pm 0.11$ &\textbf{ 4.11$\pm 0.13$}	 	 \\ \hline
{\bf USPS 4 }&  2.75$\pm 0.28$	& \textbf{3.83$\pm 0.04$}	& 3.75$\pm 0.06$	& 3.78$\pm 0.12$	 & 3.81$\pm 0.10$	 	 \\ \hline
{\bf MNIST 5 }&  3.71$\pm 0.25$	& 5.95$\pm 0.07$	& 5.75$\pm 0.10$	& \textbf{6.31$\pm 0.09$}& 6.26$\pm 0.08$	 	 \\ \hline
{\bf MNIST 6 }&  2.91$\pm 1.06$	& 6.12$\pm 0.31$	& 5.51$\pm 0.27$	& 7.02$\pm 0.19$ & \textbf{7.16$\pm 0.17$}	 	 \\ \hline
{\bf ISOLET 1 }&  3.75$\pm 1.51$	& 5.48$\pm 0.22$	& 4.95$\pm 0.29$	& 6.91$\pm 0.18$ & \textbf{7.13$\pm 0.21$}	 	 \\ \hline
{\bf ISOLET 2 }&  3.66$\pm 1.12$	& 5.04$\pm 0.20$	& 4.85$\pm 0.34$	& \textbf{7.96$\pm 0.25$} & 7.51$\pm 0.23$	 	 \\ \hline
{\bf Caltech Face }&  5.16$\pm 1.02$	& 7.92$\pm 0.17$	& 6.53$\pm 0.12$	& \textbf{10.88$\pm 0.09$} & 8.27$\pm 0.17$	 	 \\ \hline
{\bf Caltech Car }&  4.01$\pm 1.35$	& 7.03$\pm 0.18$	& 6.25$\pm 0.11$	& \textbf{8.23$\pm 0.16$} & 7.61$\pm 0.25$	 	 \\ \hline

\end{tabular}
\end{center}
\end{table}
\normalsize
%
%
%
%
%

As can be inferred from tables, the two proposed methods perform better and yield more accurate results than other methods. This is mostly because these methods asses uncertainty of data samples and select the most uncertain.

Moreover, we can see that the margin sampling approach and entropy method both give promising results and their performance is very similar. This is because of the fact that for two-class problems, both margin and entropy yield precise measures of uncertainty in prediction and utilize all  information in the posterior distribution. However, for multi-class problems (which is out of the scope of this work) entropy outperforms margin and yields more precise uncertainty measures since it uses probability of all classes rather than just the two most likely ones.

\begin{figure}[t]
\begin{center}
\includegraphics[width=2.88in,height=2.4in]{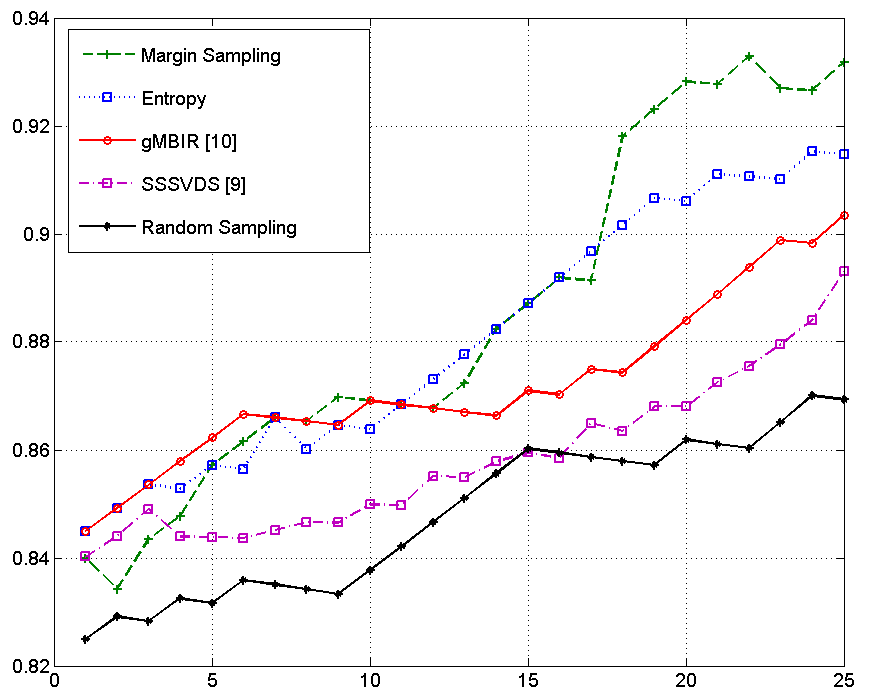}  
\caption{Plot of precision against no. of added samples for five active learning methods}
  \label{bigpo}
  \end{center}
\end{figure}

We also plotted the graph of precision against number of added samples for the Caltech Face dataset. Figure \ref{bigpo} depicts the graph for the five the active learning methods.

\begin{figure*}[t]
\begin{center}
\includegraphics[width=5in,height=0.3in]{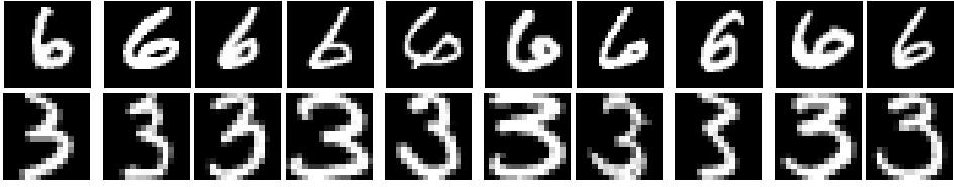}  
\caption{First 10 selected data by the margin method for USPS digit 3 (bottom) and MNIST digit 6 (top)}
  \label{sel}
  \end{center}
\end{figure*}

Figure \ref{sel} depicts top 10 selected data by the margin method for two classes of USPS and MNIST datasets. As can be seen in the figure, selected data are abnormal representatives of their corresponding class and therefore can be considered uncertain or unconfident samples which are more informative and yield better efficiency gain if their label would be known.

%


\section{Conclusion}

We proposed an active learning algorithm which utilizes only positive and unlabelled data for selecting most uncertain samples for querying in active learning. Our approach can utilize many models of one-class learning as the base classifier and the density estimation method. 

The ideas presented in this paper can be utilized in other active learning paradigms as well. For example, a very well-known and principled framework for active learning is the risk minimization approach which tries to find the data sample which yields a model with minimum possible risk on training set. The risk computation which involves both  negative and positive class posterior information, can be easily adapted to work with positive and unlabelled data by direct computation of sample likelihood and the expectation approach presented in this paper.

\section*{Acknowledgement}

The authors would like to thank the AICTC Research Center for supporting this work.

\bibliographystyle{plain}
\bibliography{refs}

\end{document}